\documentclass[conference]{IEEEtran}
\IEEEoverridecommandlockouts
\usepackage{cite}
\usepackage{amsmath,amssymb,amsfonts}
\usepackage{algorithmic}
\usepackage{graphicx}
\usepackage{textcomp}
\usepackage{xcolor}
\usepackage{multirow}
\usepackage{url}
\usepackage{comment}
\def\BibTeX{{\rm B\kern-.05em{\sc i\kern-.025em b}\kern-.08em
    T\kern-.1667em\lower.7ex\hbox{E}\kern-.125emX}}
\usepackage{amsmath,amssymb,amsfonts}
\usepackage[mathscr]{euscript}
\usepackage{makecell}
\usepackage[%
    colorlinks=true,
    pdfborder={0 0 0},
    linkcolor=blue
]{hyperref}

\renewcommand{\figurename}{Figure}

\begin{document}

\title{Auto Detecting Cognitive Events Using Machine Learning on Pupillary Data}

\author{\IEEEauthorblockN{1\textsuperscript{st} Quang Dang}
\IEEEauthorblockA{\textit{Computer Science and Electrical Engineering } \\
\textit{University of Maryland, Baltimore County}\\
Baltimore, United States \\
qdang1@umbc.edu}
\and
\IEEEauthorblockN{2\textsuperscript{nd} Murat Kucukosmanoglu}
\IEEEauthorblockA{\textit{D-Prime LLC}\\
Washington, D.C., United States \\
murat.kucukosmanoglu@dprime.ai}
\and
\IEEEauthorblockN{3\textsuperscript{rd} Michael Anoruo}
\IEEEauthorblockA{\textit{Computer Science and Electrical Engineering } \\
\textit{University of Maryland, Baltimore County}\\
Baltimore, United States \\
manoruo1@umbc.edu}
\and
\IEEEauthorblockN{4\textsuperscript{th} Golshan Kargosha}
\IEEEauthorblockA{\textit{Computer Science and Electrical Engineering } \\
\textit{University of Maryland, Baltimore County}\\
\textit{D-Prime LLC}\\
Washington, D.C., United States \\
golshak1@umbc.edu}
\and
\IEEEauthorblockN{5\textsuperscript{th} Sarah Conklin}
\IEEEauthorblockA{\textit{Computer Science and Electrical Engineering } \\
\textit{University of Maryland, Baltimore County}\\
\textit{D-Prime LLC}\\
Washington, D.C., United States \\
sarahc7@umbc.edu}
\and
\IEEEauthorblockN{6\textsuperscript{th} Justin Brooks}
\IEEEauthorblockA{\textit{Computer Science and Electrical Engineering } \\
\textit{University of Maryland, Baltimore County}\\
\textit{D-Prime LLC}\\
Washington, D.C., United States \\
jbrook1@umbc.edu}
}

\maketitle

\begin{abstract}

Assessing cognitive workload is crucial for human performance as it affects information processing, decision-making, and task execution. Pupil size is a valuable indicator of cognitive workload, reflecting changes in attention and arousal governed by the autonomic nervous system. Cognitive events are closely linked to cognitive workload as they activate mental processes and trigger cognitive responses. This study explores the potential of using machine learning to automatically detect cognitive events experienced using individuals. We framed the problem as a binary classification task, focusing on detecting stimulus onset across four cognitive tasks using CNN models and 1-second pupillary data. The results, measured by Matthew’s correlation coefficient, ranged from 0.47 to 0.80, depending on the cognitive task. This paper discusses the trade-offs between generalization and specialization, model behavior when encountering unseen stimulus onset times, structural variances among cognitive tasks, factors influencing model predictions, and real-time simulation. These findings highlight the potential of machine learning techniques in detecting cognitive events based on pupil and eye movement responses, contributing to advancements in personalized learning and optimizing neurocognitive workload management.

\end{abstract}

\section{Introduction and Related Work}\label{sec:into}

\par Cognitive workload, the mental effort or cognitive demand associated with a specific task, constitutes a fundamental concept in human cognitive neuroscience. Interdisciplinary efforts to quantify cognitive workload in real time are of widespread interest with broad applications. It is a cornerstone concept in human cognitive neuroscience with widespread interest and diverse applications \cite{Chikhi_2022}. Understanding an individual's cognitive workload provides insight into their mental state and cognitive status, facilitating personalized learning and real-time cognitive assessments. Innovative analytic tools and approaches such as machine learning have emerged for analyzing and estimating levels of cognitive workload from physiological signals obtained with electrophysiology \cite{gupta_subject-specific_2021}, electrocardiography \cite{eilebrecht_relevance_2012}\cite{oreilly_computational_2012}, electrodermal activity \cite{kosch_your_2019}, and eye tracking \cite{klingner_measuring_2008}.

\par Eye tracking, a non-invasive measure of real-time gaze patterns, pupil dilation, and saccadic movements, offers valuable insights into the dynamic functioning of the human brain \cite{holmqvist_eye-tracking_2017}. The task-evoked pupillary response (TEPR) is a change in pupil diameter that occurs in response to cognitive loading. Evidence suggests that the TEPR reflects the amount of cognitive effort required to perform the task. TEPR is observed during widespread cognitive processes \cite{van_der_wel_pupil_2018} and has been proposed to function as both a gauge and a filter for optimizing cognitive functioning \cite{ebitz_both_2018}. TEPR focuses on attention \cite{Unsworth_Pupillary}, auditory discrimination \cite{baldock_task-evoked_2019}, mathematical problem solving, visual working memory, long-term memory tasks \cite{hoffing_dissociable_2020}, multiple object tracking \cite{alnaes_pupil_2014}, decision-making \cite{lempert_relating_2015}, naming tasks \cite{loo_individual_nodate}, and vigilance \cite{martin_pupillometry_2022}. Dynamic changes in TEPR serve as essential tools for understanding cognitive processes, individual differences in cognitive abilities, resource allocation, and preferences \cite{beatty_task-evoked_1982}.

\par While multiple researchers classify cognitive workload using machine learning \cite{GUPTA2021103070} \cite{lobo2016cognitive} \cite{Blanco_2018}, there is no consensus on an agreed-upon method for quantifying it. Results can vary depending on the dataset and methodology used. For example, Gupta et al. introduces a method for estimating cognitive workload using EEG data, functional brain connectivity, and deep learning algorithms \cite{gupta_subject-specific_2021}. Their approach achieves an classification accuracy of 80.87\%, categorizing cognitive workload into low, medium, and high levels using EEG data.

\par The selection of pupillary and eye movement data as our primary focus is due to its non-invasive collection method. While multiple studies highlight the significance of the TEPR in measuring cognitive workload \cite{klingner_measuring_2008} \cite{rafiqi_pupilware_2015} \cite{Pfleging_Model}, it remains an open question whether TEPR and other pupillary features can identify the occurrence of cognitively-loading events. Based on the TEPR model, the changes in pupil diameter should indicate when a stimulus occurs and its cognitive load. Additionally, it is unclear if pupillary responses are consistent across different cognitive functions. We try to answer the central research questions whether the pupillary response dynamics immediately after stimulus onset can accurately detect the stimulus onset time, and can this response generalize across all cognitive domains.

\par Drawing inspiration from Gupta et al., our goal is to construct CNN models for predicting stimulus onset associated with TEPR using exclusively pupillary data. Unlike Gupta et al., who develop a variety of ML models, we focus on building baseline CNN models. Instead of conducting widespread multiple model testing, our approach conducts an in-depth analysis of comparing the performance of generalized versus specialized models and the factors influencing ML predictions

\par Aside from Gupta, multiple researchers have attempted to use machine learning-based approaches to analyze pupillary data for predicting cognitive workload in TEPR \cite{SHARMA2021104589}. Some researchers have achieved an accuracy of 97\% \cite{10555690}. However, few have focused on predicting cognitive events. There are advantages to predicting cognitive events over cognitive workload. Cognitive workload requires time to fully manifest. The model typically predict "low" or "high" states, which can be debatable to quantify cognitive workload. In contrast, predicting the cognitive events is the same as predicting external load. It will require shorter data segments, and we have ground truth for where the cognitive events occur. In the TEPR, the cognitve events are the stimulus onset time.

\par In our study, we build machine learning models that take pupillary and gaze data as inputs. The data is from four TEPR tasks that sample four cognitive domains: vigilance, emotion processing, numerical processing, and short-term memory. We construct CNN models that can predict stimulus onset times, using 1-second segments of pupillary data without the need for additional physiological measurements such as ECG or EEG.

\par In this paper, we have four main objectives to explore. Firstly, we construct five machine learning models: four task-specific models corresponding to four cognitive tasks, and one generalized model trained on the entire dataset. We evaluate the results for these five models and analyze the trade-off between generalization and specialization. Through this analysis, our goal is to identify common patterns that could improve the generalization capabilities of the models.

\par Secondly, we analyze cross-domain results to evaluate how the four task-specific models perform when encountering cognitive events in tasks they were not trained on. This analysis helps determine whether cognitive events can generalize across multiple domains or if they are dependent on their specific cognitive domain.

\par Thirdly, we analyze the factors influencing the model's predictions. There might be multiple factors contributing to the model's decision-making process. Our goal is to ensure that the primary contributor to the model's predictions will be human cognitive reactions. For example, if participants fixate on specific points on the screen when cognitive events are present, the model might rely on these signals for prediction rather than focusing on human cognitive reactions. This could undermine the objectives of our study.

\par Lastly, we evaluate model performance in an online environment to assess its usability in real-time scenarios. This evaluation is essential for tasks like analyzing the cognitive load of professional e-sport players during gameplay, which can only be done online. Analyzing the trade-offs between online and offline environments allows us to understand the usability and limitations of our models in practical settings.

\section{Material}\label{sec:material}

\par The dataset was part of the Cognitive Resilience and Sleep History research project. Fifty-seven participants volunteered in the research experiment. The University of California Santa Barbara Human Participants Committee (\#IRB00000307) and the Army Research Laboratory Human Research Protections Office approved all study procedures. All participants provided informed written consent to participate in the study. Each participant followed a standardized protocol, beginning with a relaxation period lasting six minutes (referred to as "REST"), followed by the completion of four distinct cognitive tasks: the Psychomotor Vigilance Task, Dot Probe Task, Mental Arithmetic, and Visual Working Memory. All participants were requested to return for subsequent sessions, with the opportunity to repeat the experiment up to ten times on separate days. The range of retesting sessions spanned from a minimum of one to a maximum of ten, yielding a median of six sessions with a standard deviation of 3.33.

\par The battery of cognitive tasks included four widely recognized cognitive assessments designed to probe various facets of mental functioning. Herein, we provide a brief overview of each task:

\par \textbf{Psychomotor Vigilance Task (PVT):} Participants were instructed to respond by pressing a designated key upon the appearance of a visual stimulus on the screen. The PVT was a classic measure of sustained attention that measures how participants respond to a simple visual stimulus for an extended period of time. Each experimental session consisted of 77 trials.

\par \textbf{Dot Probe Task (DPT):} This task involved the simultaneous display of two facial images, each categorized by emotion (angry, happy, or neutral). A subsequent visual probe would appear, and participants were asked to identify the probe's location (left or right).The DPT measured how much faster participants respond to angry stimuli compared to neutral stimuli. This is a classic assessment of selective attention. A total of 160 trials were presented in each experimental session.

\par \textbf{Mental Arithmetic (MA):} Participants were engaged in modular arithmetic problems, which varied in difficulty level (easy and hard). MA tasks were considered a core component of human logical thinking and relate to attention, working memory, processing speed, MA ability, and executive function. The difficulty level was presented randomly. Each experimental session consisted of 40 trials.

\par \textbf{Visual Working Memory (VWM):} VWM allowed for the retention of visual information to be held in the mind after the stimulus has disappeared. This task required participants to memorize a stimulus image, followed by the presentation of a second image. The second image could be similar or different from the stimulus image. Participants were asked to determine whether the second image matched the first, with difficulty levels varying (1 item in the easy level and 6 items in the hard level, presented randomly). Each experimental session consisted of 48 trials.

\par Data acquisition during the experiments involved recording various parameters at a sampling rate of 250 Hz, including Pupil Diameter (PD) and gaze position (Gaze X and Gaze Y). Additionally, behavior data were collected, encompassing the initiation time of each trial (Stimulus Time, ST), the participants' response times (Response Time), the accuracy of their responses, and task-specific information. All behavior data was documented for comprehensive analysis. In this paper, ST was used as a primary predictor of cognitive events.

\section{Exploratory Data Analysis}\label{sec:pupil}
\subsection{Pupil Response Analysis}\label{sec:pupil_EDA}

\begin{figure}[htbp]
\centerline{\includegraphics[width=\linewidth]{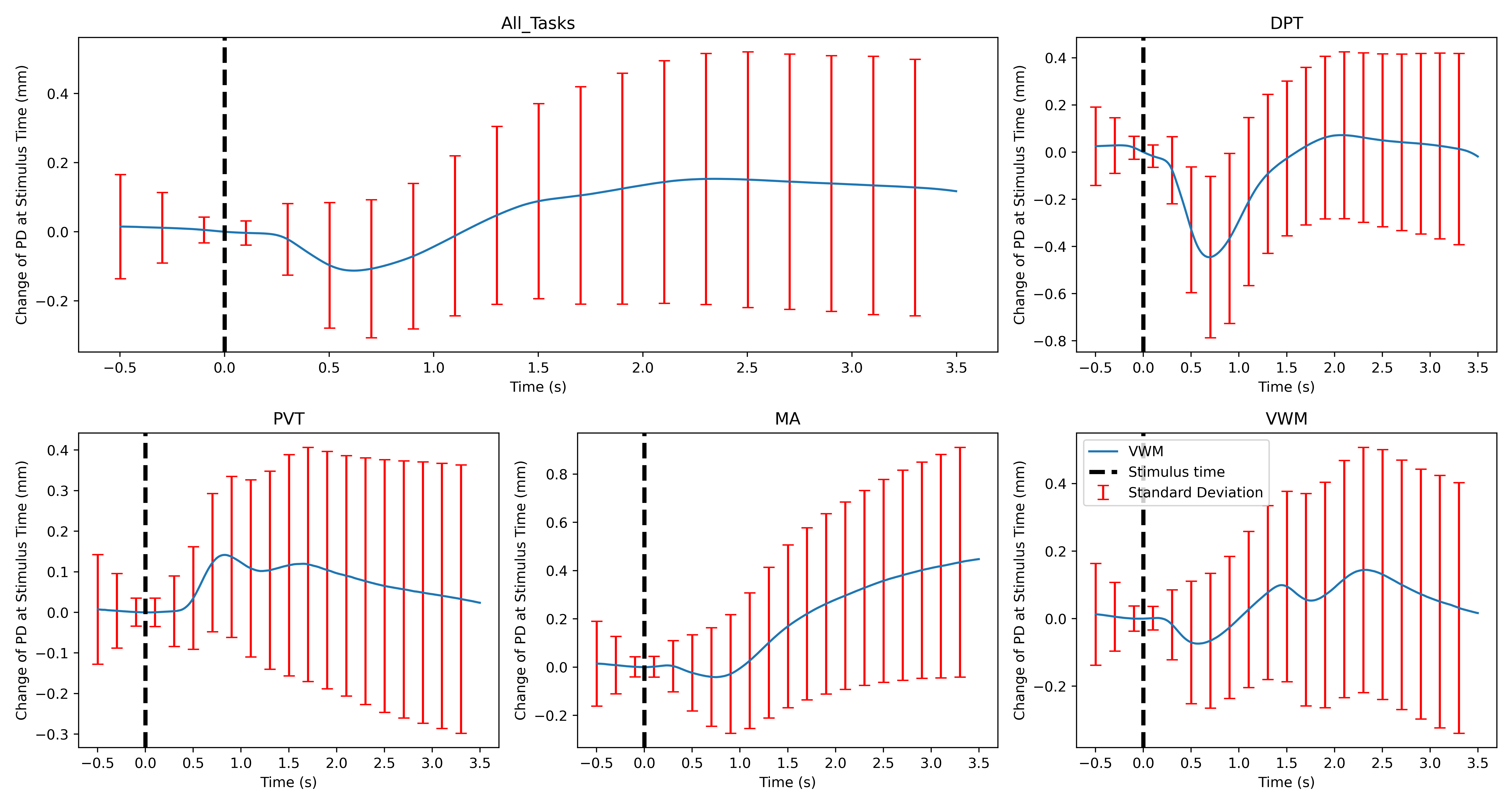}}
\caption{Mean Pupil Changed after ST for each task. Mean Change with Error Bars (1 Standard Deviation) following the start time for all 57 participants. The plots for DPT, PVT, MA, and VWM represented the respective task averages, while the ``All-tasks'' plot indicated the overall average across all datasets. The X-axis depicted time relative to the ST, while the Y-axis represented the difference between PD at the ST and PD at corresponding times.}
\label{fig1}
\end{figure}

\par In the initial phase of our exploratory data analysis, we focused on the common patterns in TEPR. We combined pupil data from all participants for each specific task and computed the average pupil response. The analysis window lasted for 4 seconds, from 0.5s to 3.5s after the ST, resulting in a total of 4 seconds of interest.  This time range was chosen to avoid pupillary light reflex (PLR) influences on the TEPR. We normalized all PD values by subtracting the PD at ST, resulting in the change in PD, as illustrated in \autoref{fig1}. The resulting averages were grouped by task, producing four panels representing each distinct task. The ``All-task'' panel incorporated data from all tasks and participants in the study of 57 subjects.

\par In the ``All-task'' panel, the PD increased 0.013 mm after ST, indicating minimal change in PD within this 4-second window. However, the standard deviation of 0.34 mm was high. As we moved further in the next section, this high variance was primarily due to individual differences in participants' responses to the ST, rather than significant fluctuations within the 4-second windows.

\par The PD did not show a significant changed in this 4-second window. However, examining smaller time intervals revealed some meaningful trends. From 0 to 1 second after ST, the average PD decreased from 0 to -0.12 mm, indicating pupil constriction. Then, between 1 to 2 seconds after ST, the average PD increased from -0.12 mm to 0.03 mm, signaling the dilation phase. The difference between these intervals was approximately 0.15 mm. This trend continued, with the average PD between 2 to 3 seconds after ST reaching 0.10 mm, an increase from 0.03 mm, indicating ongoing dilation.

\par From this observation, we identified that the period from 0 to 1 second after ST was the pupil constriction phase which could be due to saccadic eye movements as participants shifted their focus to a new probe on the screen or PLR related activity. The period between 1 to 2 seconds marked the re-dilation phase, where the pupil dilated back to the baseline before ST. The period after 2 seconds  showed participants experiencing cognitive workload from the ST.

\par The trend for each task slightly differed from the general trend. In the DPT, MA, and VWM tasks, we observed initial pupil constriction shortly after ST, lasting approximately 0.6 to 0.7 seconds. In contrast, the PVT task did not exhibit this constriction; instead, PD remained stable for about 0.5 seconds after ST. The DPT task showed a more significant constriction compared to the MA and VWM tasks, resulting in a longer re-dilation phase for DPT. In the MA and VWM tasks, the PD returned to baseline around 1 second after ST, while in DPT, the re-dilation phase took more than 1.5 seconds. After the re-dilation phase, we observed subsequent dilation in PD across all tasks, with the peak dilation surpassing the baseline at ST. The timing of peak dilation varied among tasks. This indicated that all tasks increased participants' cognitive workload.

\par
\begin{figure}[htbp]
\centerline{\includegraphics[width=\linewidth]{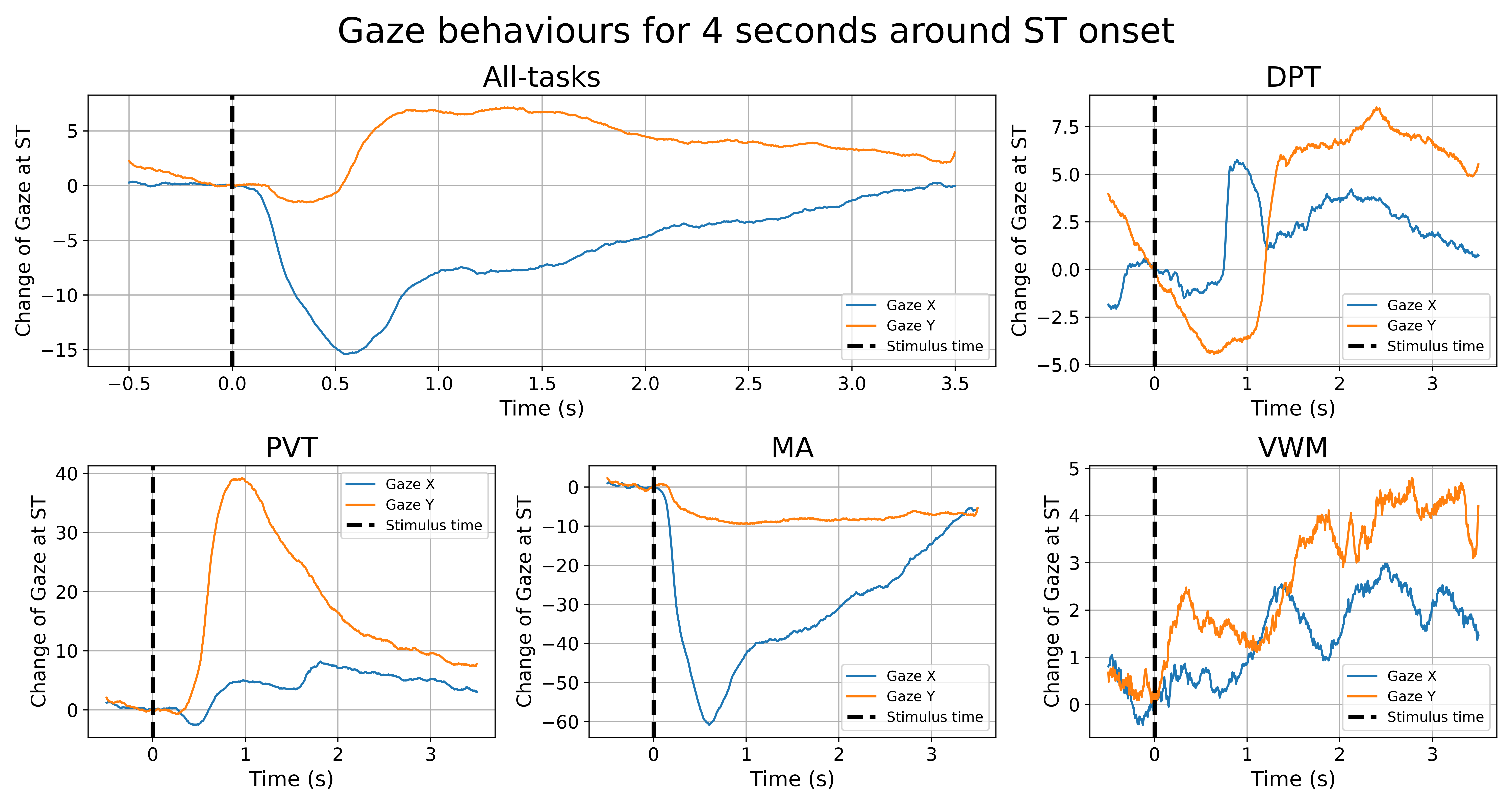}}
\caption{Mean Gaze Position Shift after ST in 57 participants for each task. The graphs for DPT, PVT, MA, and VWM showed task-specific means, while the ``All-tasks'' plot illustrated the average across all tasks. The X-axis denoted time relative to the ST, and the Y-axis displayed the variation between PD at the ST and PD at corresponding time points.}
\label{fig1.5}
\end{figure}

 The \autoref{fig1.5} showed the average gaze position data for all participants. Standard deviations were not displayed because their magnitude would have made this plot difficult to interpret. The mean for Gaze X was -1.53 and Gaze Y was 4.18; the mean standard deviation for Gaze X was 77 and for Gaze Y was 60. Unlike PD, identifying a consistent pattern for gaze position was challenging.

\par The behavior observed in the ``All-tasks'' and MA tasks showed similarity. Both tasks displayed saccadic eye movements occurring shortly after ST, lasting roughly until 0.6 seconds, followed by slower eye movements in the opposite direction. However, the PVT tasks demonstrated an inverse pattern. Initially, there was relative fixation observed from 0 to 0.4 seconds, succeeded by saccadic eye movements post the 0.4-second mark. In the case of DPT tasks, eye movements in the opposite direction were notable both before and after the 0.5-second period. However, the speed of these movements in DPT tasks appeared slower compared to MA and PVT tasks. 

\par While examining the VWM dataset, we observed high-frequency signals even after averaging. This indicated substantial noises. Averaging reduced high-frequency signals within individual samples. Upon further investigation, the high frequencies were not due to noise but because of the multiple patterns which were caused by the structure of the VWM tasks. Unlike the other tasks where participants focused on specific location after ST (e.g., a dot for PVT tasks, two images for DPT tasks, and mathematical expressions for MA tasks),VWM tasks had varying difficulty levels, ranging from easy (one image) to hard (six images). This variation in task difficulty led to different levels of displayed content and points of interest. Consequently, This resulted in significant variance in gaze position and no common pattern.

\subsection{Exploring Variability in Individual Pupil Diameters}\label{sec:pupil_vari}

\par PD was a physiological measure influenced by various factors such as individual biology, lighting conditions, and stress levels. On average, PD ranged around 4mm, but it could vary significantly among individuals \cite{spector_pupils_1990} \cite{joshi_pupil_2020}.

\begin{figure}[htbp]
\centerline{\includegraphics[width=\linewidth]{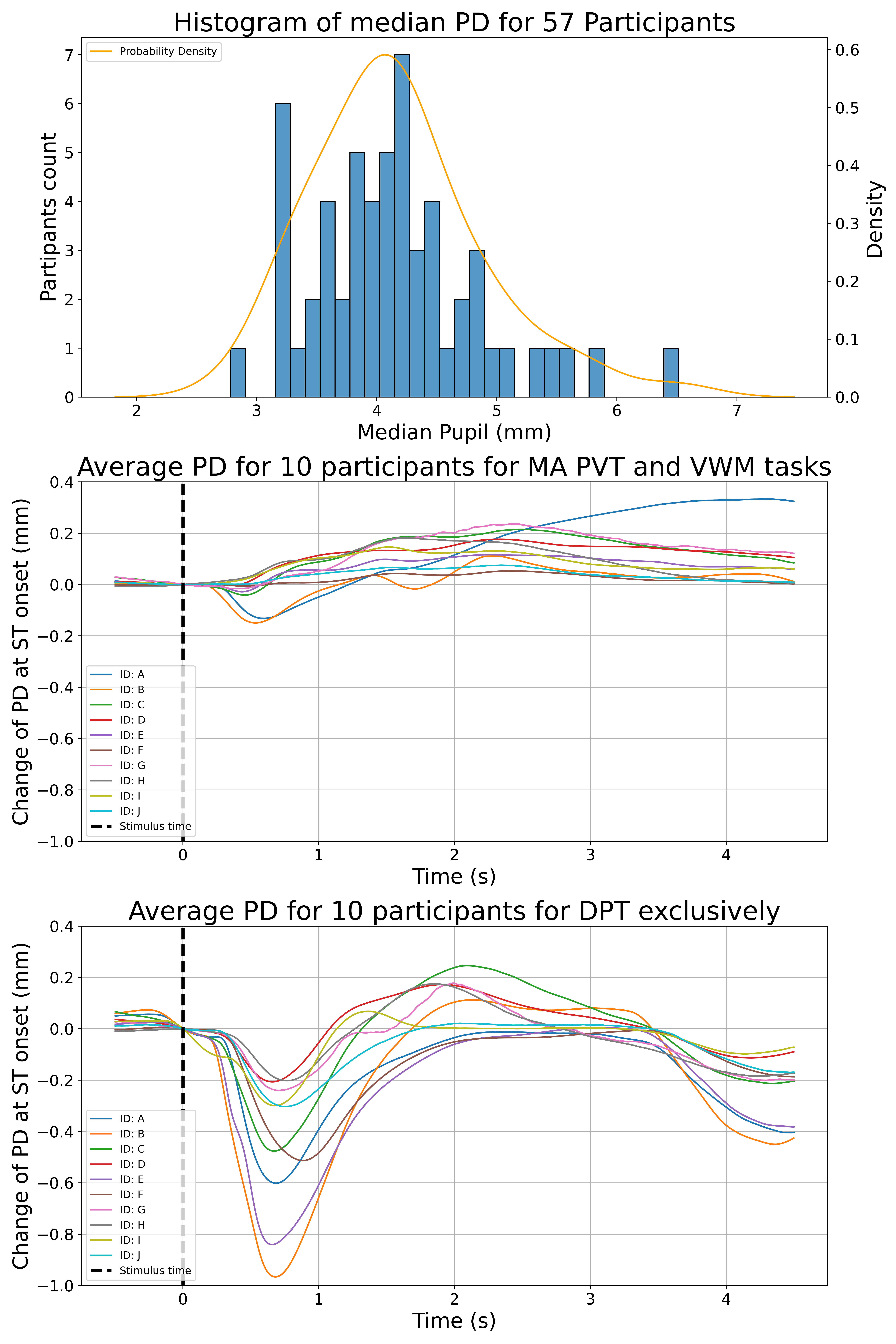}}
\caption{Analysis of Individual Participant Pupil Diameters. The top panel presented the distribution of median pupil diameters across 57 participants. Participant count was represented on the left Y-axis via bars, while the probability density function was depicted on the right Y-axis (yellow line). The X-axis represented median Pupil Diameter in millimeters. The middle and lower panels displayed the mean changes in pupil diameter over time in response to a stimulus for each participant. The X-axis indicated time relative to the ST, while the Y-axis represented the variation between pupil diameter at the ST and pupil diameter at corresponding time points. The middle panel showed the average of three tasks: MA, PVT and VWM, while the lower panel focused exclusively on the DPT.}
\label{fig2}
\end{figure}

\par To illustrate the range of PD values within our dataset, \autoref{fig2} (top panel) presented a histogram of the median PD measurements across all tasks for all fifty-seven participants. The majority of participants showed median PD values falling within the range of 3.5 to 4.5 mm. There were notable outliers, with some participants demonstrating median PD measurements exceeding 5.5mm, while others displayed medians below 3 mm. This wide variability in PD emphasizes the necessity of considering individual differences when analyzing PD data.

\par Moving on to analyze participant responded to the ST, in \autoref{fig2} (middle and lower panel), we examined the PD changes following the stimulus time for a subset of 10 participants. For visualization purposes, we limited the display to the 10 participants who participated most frequently for repeated experiments. The computation method aligned with \autoref{fig1}; however, in this case, we calculated averages on a participant-specific rather than task-specific. 

\par The \autoref{fig2} (middle panel) focused on computing the averages across three tasks: PVT, MA, and VWM, excluding the DPT task. In the \autoref{fig2} (middle panel), the participant's A and B showed the most notable pupil constriction in response to trial onset, while the participants H and I did not display significant pupil constriction. Meanwhile, the participants F and J exhibited minimal PD fluctuations, whereas Participant A had the most rapid changes. This diversity in pupil behavior underscored its significance and was a contributing factor to the large standard deviation seen in \autoref{fig1}. 

\par The \autoref{fig2} (lower panel) showed the average PD for DPT task exclusively. During our analysis, we observed a higher drop in PD during the DPT task than any other tasks. Further investigation revealed that this drop could be attributed to the pupillary light reflex (PLR). In the DPT task, the screen displayed before the ST was relatively dim compared to when the ST was presented, resulting in a sudden increase in brightness. Therefore, the majority of PD constriction observed during the DPT task was due to PLR rather than a cognitive event. This distinction was crucial as it could potentially mislead the analysis. Additionally, the DPT task contributed a substantial portion of the dataset, as previously mentioned in Section \ref{sec:material}. Each experimental session involved 160 trials in the PDT task, while the PVT, MA, and VWM tasks consisted of 77, 40, and 48 trials, respectively. Given the substantial impact of PLR and the dominance of the DPT task in the dataset, we made the decision to present the DPT task separately to accurately reflect these findings. 

\par In the \autoref{fig2} (lower panel), which specifically examined the DPT task, a common pattern showed: a constriction phase within the first 0.6 - 0.7 seconds after ST presentation. However, the magnitude varied among participants, reflecting differences in sensitivity to PLR. Notably, Participant E showed significant constriction only in the DPT task and not in the other three tasks, highlighting individualized PD responded to cognitive workload. This underscored the importance of considering unique pupil behavior between each participant.

\section{Methodology}\label{sec:method}
\subsection{Defining the Machine Learning Problem}\label{sec:preprocessing}

\par Our goal was to auto-detect cognitive events. ST signified the beginning of each trial, marking new information for participants to process. Therefore, each ST was a cognitive event, and we wanted to predict the location of ST. The input data had one-second intervals of three time-series features, Pupil Diameter (PD) and the 2D Gaze position (Gaze X and Gaze Y), each recorded at a sampling rate of 250 Hz. The problem was framed as a binary classification task, where the model's output yielded a probability ranging from 0 (indicating the absence of ST) to 1 (indicating the presence of ST). To standardize our dataset, we used standard scalers. To prevent data leakage, we segmented the standardization process, conducting it separately for each feature, each participant, each session, and each task. A single z-score was calculated by the time-series data of a single feature within one task, one participant, and one session.

\par For every ST occurrence, three samples were generated: two ``0'' samples and one ``1'' sample. Samples spanning from 0.5 seconds before ST to 0.5 seconds after ST were labeled as ``0'', samples from 0.5 seconds to 1.5 seconds after ST were labeled as ``1,'' and samples from 1.5 seconds to 2.5 seconds after ST were labeled as ``0''. We chose the start time of 0.5 seconds to minimize the influence of the PLR that may occur with stimulus presentation \cite{bremner_pupillometric_2012}\cite{ebitz_both_2018}. Typically, the initial constriction phase of the pupil took around 0.9s to 1s \cite{belliveau2023pupillary}\cite{carrick2021pupillary}. However, in our dataset, we found that the pupil reaches its minimum diameter within approximately 0.6 to 0.7 seconds. Since we aimed to capture cognitive reactions rather than PLR reactions, we chose to avoid the majority of the constriction phase. We included the bottom of the PD in our samples as it could aid in predicting the onset of cognitive processes. From \autoref{fig1}, the pupil dilation happened after 0.5 seconds. Therefore, we determined that a 0.5 to 1.5-second window time-frame was appropriate for predicting cognitive events.

\par
\begin{figure}[htbp]
\centerline{\includegraphics[width=\linewidth]{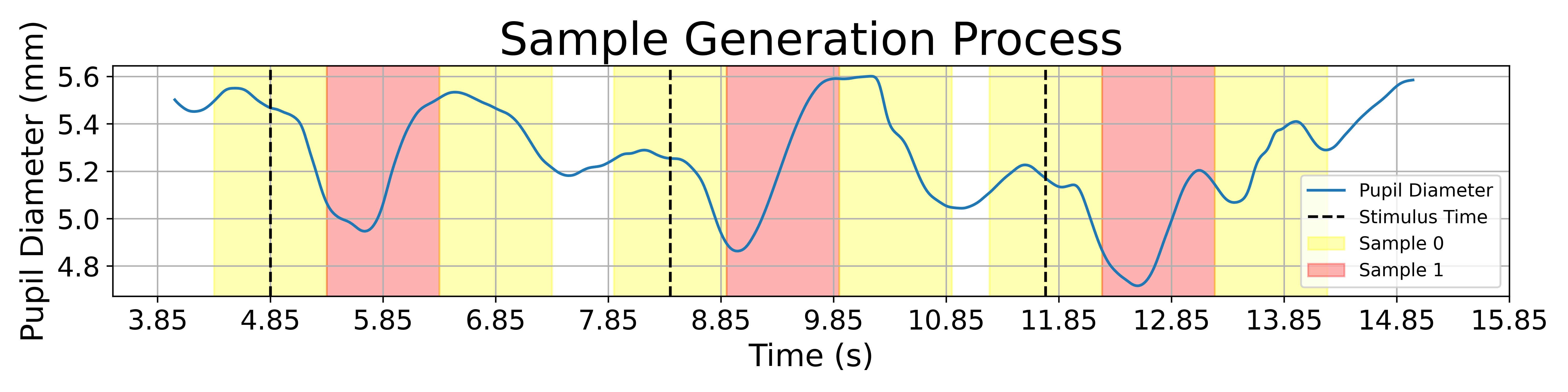}}
\caption{Sample Generation Process. The plot illustrated the process of sample generation. The black dotted line represented the Stimulus Time (ST). For every ST, the generation process produced three samples: two ``0'' samples (yellow) and one ``1'' sample (red). Each sample contained one-second of data.}
\label{fig3}
\end{figure}

 The sample from 1.5 to 2.5 seconds after ST labeled as ``0''. From the previous analysis, this time frame was where the participants continued to increase their cognitive workload. Since our goal was to detect cognitive at ST, not the manifestation of cognitive workload, we labeled this as ``0'' to help the model differentiate the PD behaviors between at ST and further from ST. The process of sample generation was illustrated \autoref{fig3}.

\par Following data sampling, any instances of missing data samples were removed. The dataset exhibited a slight imbalance, with a ratio of approximately 2:1 in favor of the ``0'' class samples. However, we chose not to address this imbalance and our subsequent results demonstrated a robust tolerance towards class imbalance. The dataset had 57 participants. 47 participants allocated as training set for training the model and the remaining 10 participants would use as testing set for testing and evaluation purposes.

\subsection{CNN architecture}\label{sec:CNN}
\par Our model's architecture was a sequential CNN comprising four sequential layers. Each layer included a convolutional layer, followed by a max-pooling layer, and a dropout layer. This architecture had two outputs: a classification output $\hat Y_{clf}$ and a PD output $\hat Y_{PD}$. The $\hat Y_{clf}$ generated a probability score ranging between ``0'' and ``1'', where ``0'' indicated the absence of ST, and ``1'' indicated the presence of ST. Additionally, we applied an auto-encoder technique to $\hat Y_{PD}$ to reconstruct the PD from the input sample.  The model's output could be defined as:

\begin{equation}
(\hat Y_{clf}, \hat Y_{PD}) = \mathscr{M}_{CNN}(X) \label{eq:1}
\end{equation}

where X was the input sample and $\mathscr{M}_{CNN}(X)$ was the CNN model, and $\hat Y_{PD}$ would used for review and error-checking purposes. To optimize the model's learning objectives, we used a composite loss function that included both cross-entropy loss, $\mathscr{L}_{cel}(X)$ for the classification output $\hat Y_{clf}$ and L1 loss $\mathscr{L}_{MAE}(X)$ for the PD output $\hat Y_{PD}$. The loss function could be expressed as:

\begin{equation}
\mathscr{L}(\mathscr{M}_{CNN}(X); Y) = \mathscr{L}_{cel}(\hat Y_{clf}, Y) + \alpha * \mathscr{L}_{MAE}(\hat Y_{PD}, X_{PD})\label{eq:2}
\end{equation}
where Y was the classification label (0 or 1), $\alpha = 0.004$ was a constant scalar, and $X_{PD}$ was the input sample containing only PD after the removal of Gaze X and Gaze Y.

\par We trained five distinct models, all using the same CNN architecture described earlier. The key distinction among these models was the training data employed. Four models were task-specific, namely ``DPT'', ``PVT'', ``MA'', and ``VWM'', trained exclusively on data from their respective tasks. In the opposite, the ``All-task'' model was trained on the entire available training dataset, aiming to evaluate the CNN architecture's ability to generalize across various tasks.

\subsection{Evaluation Metric and Feature Importance Algorithms}\label{sec:training_eval}

\par In the \ref{sec:Eval} section, we selected the Matthews Correlation Coefficient (MCC) as the primary metric. This choice was because of an imbalanced dataset, with a 2:1 ratio in favor of ``0'' samples. MCC was known for its robustness in handling imbalanced datasets \cite{chicco_advantages_2020} \cite{boughorbel_optimal_2017}. Alongside MCC, we included several secondary metrics such as accuracy, F1 score, and AUC.

\par For the third objective mentioned in the \ref{sec:into} section, we aimed to understand the factors influencing our model's predictions by conducting a feature importance analysis. This analysis helped identify which aspects of the data carry the most weight in predicting outcomes. We used the Permutation Feature Importance algorithm \cite{scikit-learn} to compute feature importance.

\par Initially, we established a baseline score, denoted as $s_{base}$. We evaluated testing samples without any alterations. $s_{base}$ was measured in terms of MCC. Then, we assessed the score for each feature. To accomplish this, we systematically substituted all the data within a specific feature with random numbers drawn from the same distribution as the original feature, keeping all other features unchanged. This process allowed us to isolate the impact of the feature under examination. Following this, we conducted predictions and computed the score for these perturbed samples, denoted as $s_{i}$, anticipating a decline in performance compared to $s_{base}$. The decrease in performance was quantified and subtracted from $s_{base}$. This iterative process was repeated for all features. To ensure the reliability of these feature importance metrics, we repeated these experiments 100 times $(N = 100)$ and calculated the mean and standard deviation across all features, as expressed in equation \ref{eq:3}:

\begin{equation}
I_{j} = \frac{1}{N} \sum_{i=1}^{N}{(s_{base} - s_{i,j})}\label{eq:3}
\end{equation}
where, $I_{j}$ represented the importance score for feature $j$, and $s_{i,j}$ was the MCC score for feature $j$ in the $i$-th iteration.

\section{Evaluation}\label{sec:Eval}

\subsection{Objective 1: Same-task Performance}\label{sec:model_perform}

\begin{figure}[htbp]
\centerline{\includegraphics[width=\linewidth]{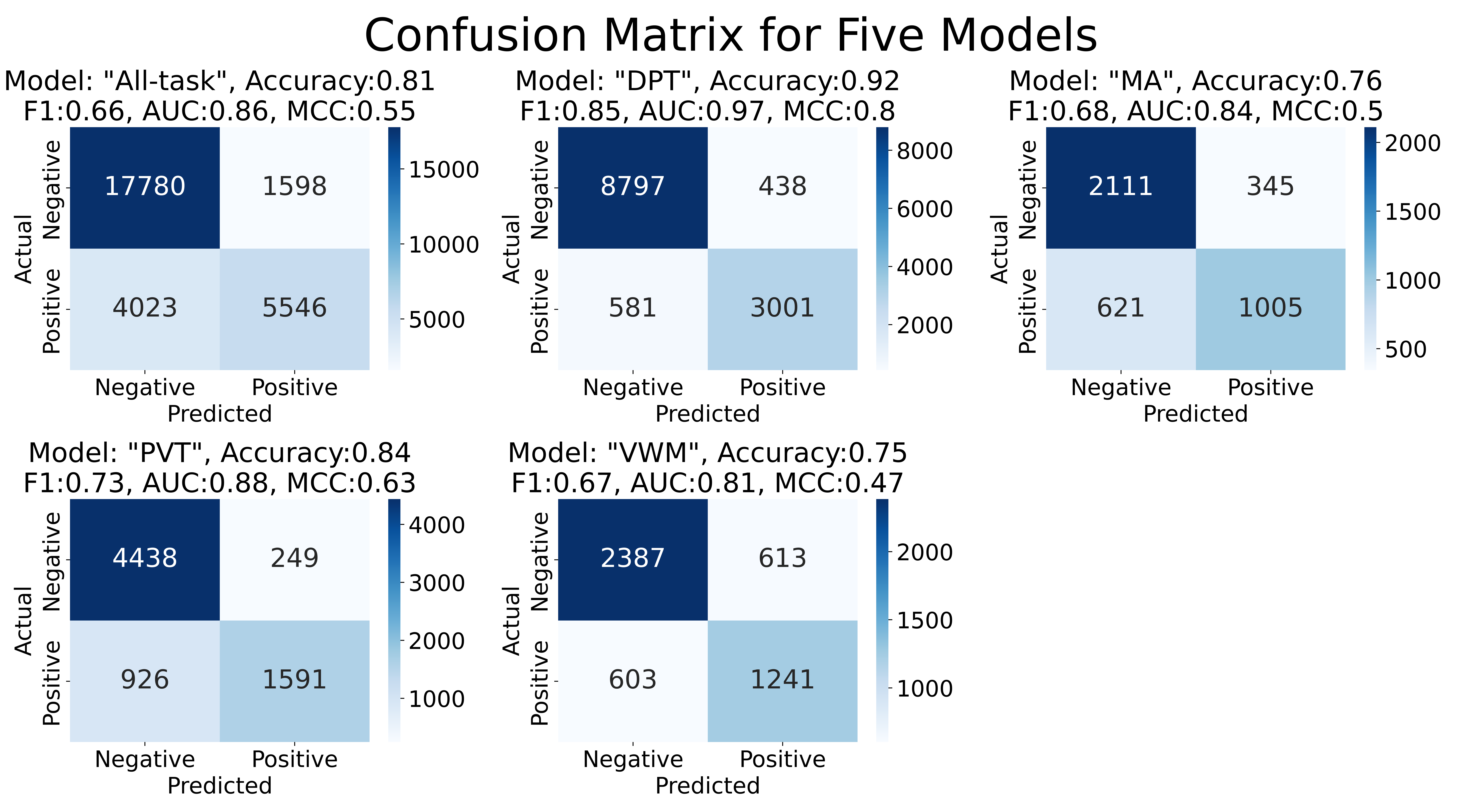}}
\caption{Confusion matrices of five models trained and tested on the same dataset. Four task-specific models (``DPT'', ``MA'', ``PVT'', ``VWM'') trained and tested separately on their respective datasets, while the ``All-task'' model trained and tested on the entire dataset. Each matrix corresponded to a single model. The panels included the ``All-task'' (upper left), ``DPT'' (upper middle), ``MA'' (upper right), ``PVT'' (lower left), and ``VWM'' (lower middle). The Y-axis represented the true label, while the X-axis represented the predictions. The model names and four metrics, namely accuracy, F1 score, AUC, and MCC, were displayed above their respective matrices.}
\label{fig7}
\end{figure}

\par We evaluated our models using a separate testing dataset containing data from 10 participants. These participants were not part of the training dataset, as previously mentioned. Task-specific models (``DPT'', ``MA'', ``PVT'', ``VWM'') were tested on their respective tasks, while the ``All-task'' model was tested on the entire dataset. The ``All-task'' model was tested on 28,947 samples. The total number of samples from all four task-specific would match the number of samples in the ``All-task'' model. The result illustrated in \autoref{fig7}

\par Across all five models, expected ``VWM'' models, the remaining four models exhibited challenges in detecting the presence of ST (``1'' samples). A consistent strength observed in four models was their ability to accurately identify when the stimulus was absent (``0'' samples), as evidenced by consistently low false positive samples compared to false negative samples. This indicated bias towards prioritizing precision over recall. In contrast, the ``VWM'' model showed a balanced performance between precision and recall, with nearly equal false positive and false negative samples.

\par Comparing the performance of the five models, measured by MCC, the ranking from worst to best was as follows: ``VWM'', ``MA'', ``All-task'', ``PVT'', and ``DPT''. However, this comparison did not highlight the trade-off between generalization and specialization because the dataset was different. Therefore, \autoref{table1} provided a direct comparison between the generalized ``All-task'' model and the four specialized models (``VWM'', ``MA'', ``PVT'', and ``DPT'') tested on the same dataset. The metrics were broken down into true positive rate (TPR) and true negative rate (TNR) to highlight the different in false positive samples and false negative samples. The red text cells indicated same-task performance where the training and testing datasets were from the same tasks (objective 1). The regular text cells indicated cross-task performance where the training and testing datasets were different (objective 2).

\begin{table}
  \centering
  \resizebox{\columnwidth}{!}{
  \begin{tabular}{|c|c|c|c|c|c|}
    \hline
    \multicolumn{2}{|c|}{} & \multicolumn{4}{c|}{\textbf{Dataset tested on}} \\
    \hline
    \multirow{7}{*}{\makecell{\textbf{Model} \\ \textbf{trained} \\ \textbf{on}}} & & \textbf{DPT} & \textbf{MA} & \textbf{PVT} & \textbf{VWM} \\
    \cline{2-6}
    & \textbf{``All-task''} & \makecell{\textcolor{red}{(TNR,TPR):} \\ \textcolor{red}{(0.94,0.73)}} & \textcolor{red}{(0.90,0.53)} & \textcolor{red}{(0.90,0.55)} & \textcolor{red}{(0.89,0.37)} \\
    \cline{2-6}
    & \textbf{``DPT''} & \textcolor{red}{(0.95,0.84)} & (0.87,0.47) & (0.83,0.03) & (0.83,0.33) \\
    \cline{2-6}
    & \textbf{``MA''} & (0.94,0.17) & \textcolor{red}{(0.86,0.62)} & (0.84,0.04) & (0.91,0.12) \\
    \cline{2-6}
    & \textbf{``PVT''} & (0.87,0.04) & (0.94,0.03) & \textcolor{red}{(0.95,0.63)} & (0.93,0.09) \\
    \cline{2-6}
    & \textbf{``VWM''} & (0.75,0.50) & (0.73,0.38) & (0.69,0.37) & \textcolor{red}{(0.80,0.67)} \\
    \hline
  \end{tabular}
  }
  \caption{In this table, the columns indicated the dataset being tested, while the rows represented the model used for testing. Each cell displayed the metrics as (TNR, TPR), with the first number being TNR and the second number being TPR. The red cells indicated instances where both the training and testing datasets were the same (same-task, objective 1). The regular cells showed instances where the training and testing datasets were different (cross-task, objective 2).}
  \label{table1}
\end{table}

\par Comparing the red cells in \autoref{table1} between the ``All-task'' model and the four task-specific models, we observed that the ``All-task'' model underperformed in most measurements. The exceptions were the TNR for the MA and VWM datasets. When converted to the MCC metric, the ``All-task'' model underperformed by about 0.1 in MCC. Notably, the TPR for the VWM dataset dropped by 0.3, which was significant compared to other TPR values, which only dropped by about 0.1. We would discuss these different behaviors of the VWM dataset in the \ref{sec:discussion} section.

\subsection{Objective 2: Cross-task Performance} \label{sec:cross_valid}

\par The purpose of the cross-task analysis was to evaluate how the models performed when encountering ST they had not seen before. For examples, the ''DPT'' model trained only on the DPT dataset, but tested on the other three datasets.  The performance of the four task-specific models showed in the 12 regular text cells in \ref{table1}.

\par Looking at TNR, given the dataset was 2:1 imbalance in favor of  ``0'' sample, the baseline for random guessing in TNR was 0.67.  All four models aimed to predict 0 for ``0'' samples. The ``VWM'' model underperformed comparing to other three models. The TNR values for ``VWM'' model were 0.75, 0.73, 0.69 while the other models had TNR values higher than 0.83, with some reaching as high as 0.93 and 0.94.

\par Now we focus on TPR. Because of imbalance dataset, the baseline for random guessing in TPR was 0.33. The results were more complex to interpret. 7 out of 12 TPR values were under 0.17. This indicated the models tended to predict 0 for ``1'' samples, meaning the four models were more likely to reject ST for other datasets. Some TPR values were as low as 0.03 and 0.04, indicating very high accuracy in rejecting ST.

\par 5 out of 12 TPR values were from 0.33 to 0.50. This signaled uncertainty, as the models were close to random guessing in these cases. Notably, all three TPR values for the ``VWM'' model fell within this range. This indicated that ``VWM'' model was more likely guess randomly when encountering other dataset. This suggests a fundamental difference in the VWM dataset compared to the other three datasets. We will discuss these differences further \ref{sec:discussion} section.

\subsection{Objective 3: Factors Influencing the Predictions} \label{sec:Pupil_only}

\par We analyzed the factors influencing the model predictions by examining feature importance. The feature importance allowed people to interpret what factors the machine learning model relied on when making predictions. Since the goal was to predict cognitive reactions, the main focus should be on PD, as it reflected cognitive responses. If Gaze Position were the dominant factor, it could indicate that the model's predictions were influenced more by the task structure than by human cognitive reactions.

\begin{figure}[htbp]
\centerline{\includegraphics[width=\linewidth]{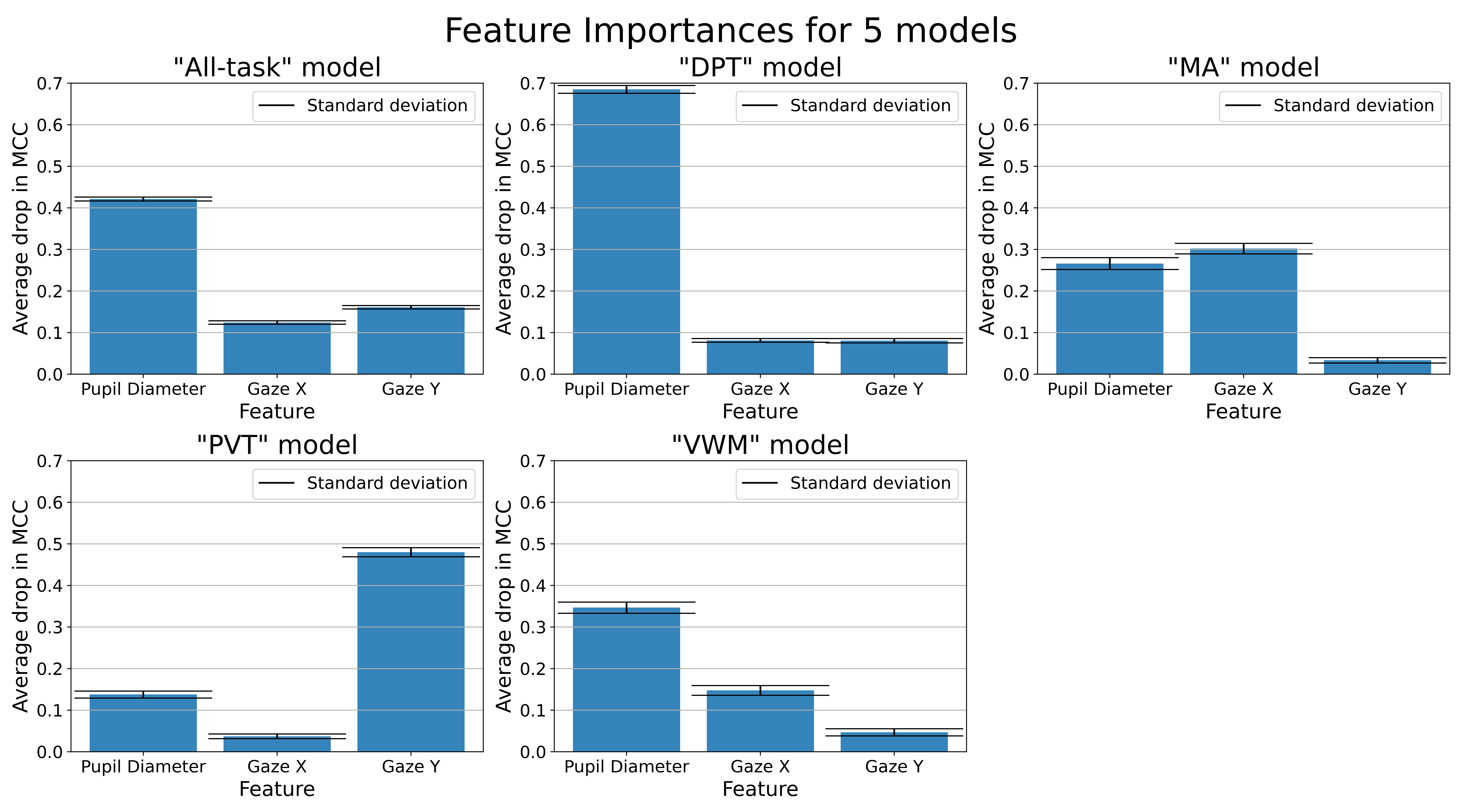}}
\caption{The figure illustrated the feature importance for five models obtained through 100 iterations of the Permutation Feature Important algorithm. Each panel represented a specific model: ``All-task'' (top left), ``DPT'' (top middle), ``MA'' (top right), ``PVT'' (lower left), and ``VWM'' (lower middle). On the X-axis were the three features used for prediction: Pupil Diameter (PD), Gaze X, and Gaze Y. The Y-axis quantified the decrease in MCC when the respective feature was distorted.}
\label{fig9.5}
\end{figure}

\par \autoref{fig9.5} provided insights into feature importance for five models, using the same calculation method mentioned in the \ref{sec:training_eval} section. From the previous analysis in \ref{sec:pupil} section, this allowed interpretation \autoref{fig9.5}. In the ``DPT'' model, PD was heavily relied upon. This initially could signal the model using cognitive reaction for its prediction. However, given the significant PLR in DPT tasks, this heavy reliance on PD might not solely reflect cognitive reactions but also PLR influences. On the other hand, the ``MA'' model showed a split between PD and Gaze X, suggesting that task structure played a role in predictions. Similarly, the ``PVT'' model indicated that Gaze Y contributed significantly to predictions, implying reliance on task structure rather than cognitive reaction. In the case of the ``VWM'' model, the substantial variance in gaze positions was due to multiple difficulty levels might render gaze position unreliable. Therefore, the model might primarily rely on PD for making predictions.

\par Interestingly, despite the task-specific models predominantly using task structure for predictions, the generalized ``All-task'' model demonstrated that two-thirds of predictions were from PD. This could indicate that incorporating a variety of tasks in training resulted in a greater emphasis on cognitive reactions for predictions.

\begin{table}
  \centering
  \resizebox{\columnwidth}{!}{
  \begin{tabular}{|c|c|c|c|c|c|}
    \hline
    \multicolumn{2}{|c|}{} & \multicolumn{4}{c|}{\textbf{Dataset tested on}} \\
    \hline
    \multirow{7}{*}{\makecell{\textbf{Model} \\ \textbf{trained} \\ \textbf{on}}} & & \textbf{DPT} & \textbf{MA} & \textbf{PVT} & \textbf{VWM} \\
    \cline{2-6}
    & \textbf{``All-task''} & \makecell{\textcolor{red}{(TNR,TPR):} \\ \textcolor{red}{(0.94,0.65)}} & \textcolor{red}{(0.88,0.41)} & \textcolor{red}{(0.87,0.30)} & \textcolor{red}{(0.86,0.36)} \\
    \cline{2-6}
    & \textbf{``DPT''} & \textcolor{red}{(0.93,0.72)} & (0.86,0.44) & (0.82,0.10) & (0.79,0.37) \\
    \cline{2-6}
    & \textbf{``MA''} & (0.95,0.44) & \textcolor{red}{(0.88,0.55)} & (0.82,0.07) & (0.85,0.33) \\
    \cline{2-6}
    & \textbf{``PVT''} & (0.90,0.09) & (0.84,0.12) & \textcolor{red}{(0.89,0.51)} & (0.89,0.14) \\
    \cline{2-6}
    & \textbf{``VWM''} & (0.90,0.26) & (0.80,0.41) & (0.87,0.17) & \textcolor{red}{(0.88,0.46)} \\
    \hline
  \end{tabular}
  }
  \caption{Using the same computational approach as in \autoref{table1}, the prediction numbers in \autoref{table2} were solely on PD, excluding Gaze X and Gaze Y. The columns indicated the dataset being tested, while the rows represented the model used for testing. Each cell displayed metrics as (TNR, TPR), with the first number was TNR and the second was TPR. Red cells indicated where both the training and testing datasets were the same (same-task, objective 1). Regular cells showed where the training and testing datasets were different (cross-task, objective 2).}
  \label{table2}
\end{table}

\par After identifying the factors influencing predictions, we evaluated how the five models performed using only PD data. The \autoref{table2} calculated using the same method as in \autoref{table1}, with the difference being that \autoref{table2} used only the PD feature, excluding Gaze X and Gaze Y, while \autoref{table1} used all three features. We compared the value between two tables.

\par Looking at same-task performance (8 red cells), all TPR and TNR values decreased. On average, the TNR dropped by about 0.05, while the TPR dropped by about 0.10. Notably, in the PVT columns, \autoref{fig9.5} indicated that Gaze Y was the dominate factor in ``PVT'' model. However, the results in \autoref{table2} were still decent and it implied the possibility to predict ST for PVT dataset using only PD. The ``PVT'' model simply preferred using Gaze Y for its prediction. The most significant decrease was the TPR of the ``All-task'' model on the PVT dataset with a drop of 0.25. This suggested that the ``All-task'' model relied on gaze position to identify ST for the PVT dataset. 

\par Looking at cross-task performance (12 regular cells), we observed that for the ``DPT'', ``MA'', ``PVT'' models (excluding the ``VWM'' model), the TNR generally decreased while the TPR increased. This implied a lower performance in rejecting ST from other datasets. On the contrary, for the ``VWM'' model, the TNR increased and TPR decreased for ``VWM'' model, indicating that gaze position negatively impacted performance for this model. If we looked at \ref{sec:pupil_EDA}, gaze position data for VWM dataset was noisy due to varying difficulty level.

\par In general, gaze position (Gaze X and Gaze Y) contributed to the prediction but was not the dominance factor. However, when training all three features, the ``MA'' and ``PVT'' model showed a preference for gaze position over PD.

\subsection{Objective 4: Real-time Simulation}\label{sec:real-time}

\par The previous results were evaluated in an offline environment, where we had access to the complete dataset, allowing us to segment and evaluate long time-series data effectively. However, in real-time scenarios, data was not all available at the beginning. The online environment required a different approach, where data became available incrementally, and models had to make predictions on the fly. In this experiment, we used the same testing dataset as in all previous sessions. To simulate the online environment, we modified the normalization process. Instead of using the entire dataset for normalization, we used the initial 60 seconds of data as a baseline, and normalized the entire dataset using that baseline, thereby preventing potential data leaking. This baseline continued to update as more data became available. For the sampling process, we used a one-second data window for making predictions, shifting every 25 data points, considering the 250 Hz data sampling rate. This corresponded to predictions every 0.1 seconds.

\begin{figure}[htbp]
\centerline{\includegraphics[width=\linewidth]{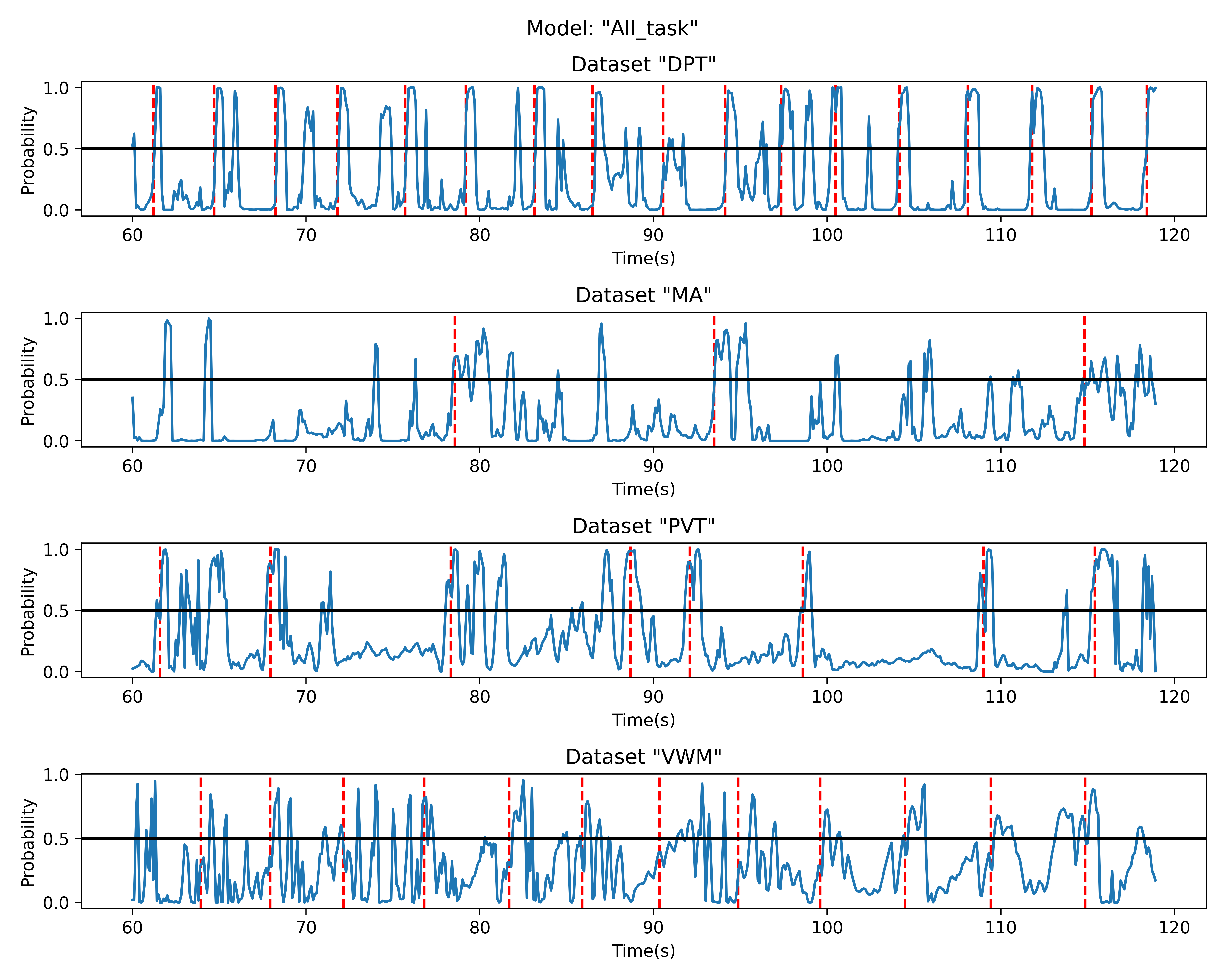}}
\caption{Real-time prediction using the ``All-task'' model on one participant. The figure showed one minute of prediction for a single participant using the ``All-task'' model. The initial 60 seconds (0-60s) reserved for data normalization. The plot showed the next 60 seconds (60-120s) for four different tasks from the same participant. The red dotted line indicated ST, while the blue line represented the probability of model predictions. The probability of 1 suggested the model predicted the presence of the ST, while the probability of 0 implied the absence of ST. Probabilities closer to 0.5 indicate that the model was unsure and more likely guessing in its predictions.}
\label{fig9}
\end{figure}

\par Displaying all results would be challenging. \autoref{fig9} showed a sample result of ``All-task'' model for 60-second output for four different datasets from one participant. The model predicted the majority of ST accurately. The most missing ST were from the ``VWM'' dataset. This highlights the ``VWM'' dataset's lower performance relative to the other three. As noted in Section \ref{sec:preprocessing}, our predictions used data from 0.5 to 1.5 seconds after the ST, introducing a 0.5-second lag. This intentional delay was maintained to assess its impact, with \autoref{fig9} showing peaks slightly delayed relative to their corresponding ST.

\par In instances where no ST, the model outputted probabilities below 0.5, indicating the absence of cognitive events. Occasionally, we observed peaks in model predictions exceeding the 0.5 threshold, suggesting potential cognitive events. While many of these instances were likely misclassifications, we could not discount the possibility of cognitive events originating outside the trial window or from participant's internal thoughts. This phenomenon was particularly noticeable in the MA dataset where, after the initial peak corresponding to the ST, we observed secondary peaks approximately 4-5 seconds later. For example, in the MA dataset panel of \autoref{fig9}, the first ST occurred around 78 seconds, with the initial peak in predictions around 78 to 80 seconds. Then, a second peak was around 82 seconds. Similar occurrences were scattered throughout the MA dataset. The first peak (ST) corresponded to the participant seeing the math problems, and the second peak might relate to the participant attempting to solve them. This suggested the possibility of unrecorded cognitive workloads outside the trial context.

\par The results for the five models in the online environment: ``All-task'', ``DPT'', ``MA'', ``PVT'', and ``VWM'', were 0.50, 0.75, 0.44, 0.58, 0.41 respectively, all measured in MCC. Compared to \autoref{fig7}, all models experienced a decrease of approximately 0.05 in MCC. This illustrates the trade-off between offline and online environments.

\section{Discussion} \label{sec:discussion}

\subsection{Key points from the results} \label{sec:result_analysis}
\par The results for five model: ``All-task'', ``DPT'', ``MA'', ``PVT'', and ``VWM'' measured in MCC were 0.55, 0.80, 0.50, 0.63 and 0.47 respectively. There were three keys topic to highlight from the results: the trade-off between generalization and specialization, the models' behaviors when encountering unseen ST, and how gaze position contributed to the prediction.  

\par \autoref{table1} provided a direct performance comparison between the general ``All-task'' model and the four task-specific models by comparing 4 red cells in ``All-task'' model with 4 red cells in the task-specific models. Notably, due to the dataset's imbalance, with a 2:1 ratio favoring ``0'' samples, the baseline for random guessing was 0.66 for TNR and 0.33 for TPR. While the TNR showed some variances, the averages of TNR were closely similar. Most of the trade-off between generalization and specialization appeared in the TPR values. The drop in TPR for the DPT, MA, and PVT datasets was about 0.11, while the drop for TPR in the VWM dataset was 0.3, a significant decline which will be discussed in the next section. In general, the trade-off between generalization ``All-task'' model and fours specialization models was about 0.1 in MCC.

\par Looking at cross-tasks (12 regular cells) in \autoref{table1}, the ``VWM'' model showed low performance, close to the random guessing baseline. The other three models, ``DPT'', ``MA'', and ``PVT'' showed high TNR and low TPR. This implied these models tended to reject ST from unseen datasets. The accuracy for some rejections was high with 0.03, 0.04 in TPR. This was equal to 97\% and 96\% accuracy in rejections.

\par The influence of gaze position could be seen by comparing \autoref{table1} (using PD, Gaze X, and Gaze Y) and \autoref{table2} (using PD only). In general, while gaze position contributed to the results, it was not the primary factor for prediction. Two key points stood out. First, the TPR for ``All-task'' model on the PVT dataset dropped from 0.55 to 0.30, indicating that the ``All-task'' model relied on gaze position to detect ST in the PVT dataset. Second, the ``VWM'' model showed improvement when gaze position was removed, likely due to inconsistent gaze patterns in the VWM dataset. However, gaze position was still necessary for predicting ST in the VWM dataset overall.

\subsection{Brain region activation differences between VWM and DPT, MA, and PVT tasks}

\par In \autoref{table1}, we observed a notable difference in the results for the ``VWM'' model compared to the other fours models. Firstly, the "VWM" model displayed a balance between precision and recall, unlike the other models, which showed a bias towards precision over recall. Second, when comparing the ``VWM'' model to the ``All-task'' model on the VWM dataset, the TNR increased from 0.80 to 0.89, while the TPR dropped significantly from 0.67 to 0.37. Considering that the baseline for random guessing in ``1'' samples is 0.33, the ``All-task'' model was mostly guessing ``1'' samples randomly. This was an unusual phenomenon. The ``VWM'' model showed a TPR of 0.67. It indicated the possibility to detect ST for the VWM dataset. However, when all four dataset were trained together, the ``All-task'' model struggled to detect ST for the VWM dataset. This discrepancy suggested a structural difference between the VWM and the remaining dataset, potentially impacting the predictability.

\par The DPT, MA, and PVT tasks engaged different aspects of attention, such as selective attention, sustained attention, and attentional disengagement. In contrast, VWM was a cognitive process that involved storing and manipulating visual information in short-term memory.

\par These attention-related tasks, like DPT, MA, and PVT, likely involved the anterior cingulate cortex (ACC), a brain region associated with various cognitive functions such as conflict monitoring \cite{braem_role_2017}, error detection \cite{orr2012error}, response selection \cite{turken_response_1999}, and executive control \cite{michelet2016opioid}. The ACC also regulated emotional processes like affective appraisal and emotional regulation, adapting attention to task demands and emotional context \cite{etkin_emotional_2011}.

\par VWM, however, might rely more on brain regions like the posterior parietal cortex \cite{pisella2017optic} and prefrontal cortex \cite{voytek_prefrontal_2010}, which are involved in encoding, maintaining, and manipulating visual information in a short-term memory buffer. The posterior parietal cortex handled spatial attention and representation \cite{szczepanski_mechanisms_2010}, while the prefrontal cortex managed executive control and working memory updating \cite{dardenne_role_2012}. VWM might not heavily involve the ACC compared to the other tasks, as it did not involve as much less conflict, error, or emotional processing.

\par It was interesting to consider whether differences in brain regions activation could affect the pupillary responses. Further testing and analysis required to delve into this question. However, the \autoref{table1} implied there might be differences. The ``All-task'' model faced challenges in detecting ST from the VWM dataset while the ``VWM'' model did not. This discrepancy could be becasue most ST instances were associated with the ACC brain region. When the entire dataset was train together, other ST could overshadow ST from the VWM dataset.

\subsection{Factor Influence the Model Prediction} \label{sec:feature_importance}

\par Detecting cognitive events using pupillary data presented significant challenges. It was not about the machine learning model's inability to detect the ST itself. The main difficulty laid in guiding the model to prioritize human cognitive reactions in its predictions. Through the \ref{sec:pupil} session and the \ref{sec:Pupil_only} session, we identified three key factors relevant to ST detection: the unique structure of tasks,PLR , and human cognitive reactions.

\par The unique task structure referred to participants directing their gaze to a specific location when the ST was presented. However, relying solely on this factor had limitations, as it tailored to specific tasks and within a lab-controlled environments. Outside of such conditions, the model's performance could under perform. In \autoref{fig9.5}, in the ``PVT'' model, Gaze Y emerged as the primary contributor, indicating a focus on specific locations rather than cognitive reactions.

\par Additionally, cognitive workload was not the sole factor in PD; the PLR also contributed to changing in PD. In \autoref{fig9.5}, the ``DPT'' model showed PD contributes to the predictions more than other tasks. This indicated the models attempting to utilize PLR for detecting cognitive events. The influence of cognitive reactions in the DPT task might be overshadowed by PLR. Separating PLR and cognitive reactions from PD could be challenging. Predicting PLR was not our objective; we wanted to predict cognitive reactions.

\par Interestingly, the ``All-task'' model predominantly relied on PD as its predictor. In contrast, the importance of PD in the ``MA'', ``PVT'', and ``VWM'' models were not as significant as in the ``All-task'' model. Our hypothesis suggested that since task-type information was not encoded into the predict samples, the ``All-task'' model relied more on cognitive reactions than task-specific patterns. While the four task-specific models depended on unique patterns within their tasks, the absence of task-type information in predicting samples prevented the model from distinguishing the difference between task types. Consequently, when trained together, these patterns might be overshadowed, and the model likely detected cognitive events based on their commonality across tasks, namely cognitive reactions. As the variety of tasks increased, we expected the ``All-task'' model to increasingly relied on cognitive reactions for its predictions. We planed to add more additional cognitive tasks to test this hypothesis.

\subsection{Pupil constriction and shifting prediction window} \label{sec:saccadic_movement}

\par The analysis in \autoref{fig1} revealed an initial pupil constriction occurring shortly after ST, lasting approximately 0.5 to 0.7 seconds, varying slightly depending on the specific task. This initial constriction could be attributed to saccadic eye movements \cite{binda_vision_2018}. After the ST, participants shifted their gaze toward new presented information, contributing to the observed constriction \cite{dicroscio2018task}. However, while saccades contribute to pupil constriction, they might not be the sole determinant. The PVT task did not exhibit significant pupil constriction, while both the MA and VWM tasks showed minimal constriction, averaging less than 0.1 mm. The DPT task showed an average pupil constriction of approximately 0.4 mm.

\par The pupil constriction in DPT tasks was primarily caused by PLR \cite{cohen_hoffing_investigating_2022} \cite{bremner_pupillometric_2012}. In the DPT task, the screen brightness varies; it was relatively dimer before the ST and became brighter when the ST was presented. This change in brightness caused greater pupil constriction in the DPT tasks compared to the other three tasks, where brightness remained unchanged. As highlighted in Section \ref{sec:preprocessing}, both PLR and saccades were  the reasons for shifting the ``1'' sample window from 0.5 to 1.5 seconds instead of from 0 to 1 second. This adjustment aimed to capture pupil dilation in response to human cognitive workload. This adjustment aimed to capture pupil dilation in response to cognitive workload, as the main focus was to analyze cognitive workload, and capturing PLR effects was not the objective.

\section*{Conclusion}
\par In this study, we developed and evaluated five CNN models using pupillary data to predict stimulus onset times for 4 different cognitive tasks. We built 1 generalized model ``All-task'' and four specialized models. The study highlighted several key findings:

\par Firstly, four task-specific models (``DPT'', ``MA'', ``PVT'', and ``VWM'') demonstrated notable performance, achieving MCC scores of 0.80, 0.49, 0.63, and 0.49, respectively. The generalized ``All-task'' model achieved an MCC of 0.54. When compared on a task-by-task basis, it underperformed the task-specific models by about 0.1 in MCC. 

\par Secondly, task-specific models tended to reject ST they were not trained on. This indicated CNN models' ability to differentiate between ST associated with different cognitive tasks. Notably, the VWM task displays different behaviours compared to the other three tasks.

\par Thirdly, four task-specific models tended to rely on unique task structures, while the ``All-task'' model leaned towards cognitive reactions. As the variety of tasks during training increases, models tended to rely more on cognitive reactions due to the overshadowing of individual task structures.

\par Lastly, during online environment stimulation, the performance of all five models decreased by approximately 0.05 in MCC. This highlighted the trade-off between a dynamic environment and data availability. 

\par Moving forward, our focus was on enhancing models by incorporating a broader range of participant-specific data. We planned integrate more information into the model without extending the sample duration. Specifically, we intended to include 180 seconds of participant baseline resting data as a baseline. This resting data, which varied among participants, could provide the model with background pupillary information specific to each participant, enhancing prediction accuracy. Furthermore, several known factors influenced PD, such as body mass \cite{SEGAL2022100417}, lighting conditions \cite{bremner_pupillometric_2012},age \cite{guillon_effects_2016}, and medication \cite{naicker_central_2016}. In future modeling efforts, we intended to incorporate these factors.

\par The implications of this research had potential applications in real-time detection of cognitively demanding tasks and accurate determination of stimulus onset times, including individual stress level identification. This research could enhance human-computer interaction, improving healthcare systems, develop personalized learning experiences, optimizing learning outcomes, and allocating workload based on individual cognitive needs.

\section*{Acknowledgment}

\par We would like to express our gratitude to Phil Beach, Mario Mendoza, Hannah Erro, and Zoe Rathbun for their contributions to data generation and study coordination. We appreciate Steven Thurman for his helpful comments. We also acknowledge the Army Research Laboratory for sponsoring this dataset. The views and conclusions contained in this document are those of the authors and should not be interpreted as representing the official policies, either expressed or implied, of the US DEVCOM Army Research Laboratory or the U.S. Government. The U.S. Government is authorized to reproduce and distribute reprints for Government purposes notwithstanding any copyright notation herein.

\bibliographystyle{plain}
\bibliography{bibiography}

\end{document}